\newcommand{\PaletteSimArxiv}{}
\documentclass[journal,10pt,twoside]{IEEEtran}
\IEEEoverridecommandlockouts
\ifdefined\PaletteSimArxiv\else
\newenvironment{IEEEkeywords}{\par\noindent\textit{Index Terms---}}{\par}
\fi

\usepackage[T1]{fontenc}
\usepackage{cite}
\usepackage{xurl}
\usepackage{graphicx}
\usepackage{wrapfig}
\usepackage{needspace}
\usepackage{stfloats}
\usepackage{amsmath,amssymb}
\usepackage{booktabs}
\usepackage{multirow}
\usepackage{xcolor}

\ifdefined\PaletteSimOverleafNotes
  \newcommand{\PaletteSimOverleafNote}[2]{%
    \par\nopagebreak\smallskip
    {\color{red!55!black}\small\raggedright
      \noindent\textbf{Overleaf / froggagul / #1.} #2\par}%
    \smallskip
  }
\else
  \newcommand{\PaletteSimOverleafNote}[2]{\ignorespaces}
\fi
\usepackage{tikz}
\usetikzlibrary{positioning,fit,arrows.meta,calc,backgrounds}
\usepackage[hidelinks]{hyperref}
\hypersetup{pdftitle={PaletteSim: Fast Multi-Engine Neural Simulation}}
\ifdefined\PaletteSimArxiv\else
\hypersetup{pdfauthor={}}
\fi

\newcommand{\palettesimsymbol}{%
  \includegraphics[height=4.8mm]{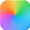}%
}

\begin{document}

\bstctlcite{palettesim_bib_style}

\title{\ifdefined\PaletteSimArxiv\else\bfseries\fi
  \texorpdfstring{\raisebox{0.15ex}{\palettesimsymbol}\hspace{0.9mm}}{}PaletteSim: Fast Multi-Engine Neural Simulation%
  \texorpdfstring{\vspace{-8pt}}{}}

\ifdefined\PaletteSimArxiv
\author{Dongwon~Son$^{1}$, Minseok~Lee$^{1}$, Hojin~Jung$^{2}$, and Beomjoon~Kim$^{1}$%
\thanks{Corresponding author: Beomjoon Kim.}%
\thanks{$^{1}$Dongwon Son, Minseok Lee, and Beomjoon Kim are with the Korea Advanced Institute of Science and Technology (KAIST), Daejeon, Republic of Korea. {\tt\small \{dongwon.son, beomjoon.kim\}@kaist.ac.kr}}%
\thanks{$^{2}$Hojin Jung is with KRAFTON. \href{mailto:hojin.jung@krafton.com}{\texttt{hojin.jung@krafton.com}}}%
}
\markboth{PALETTESIM: PREPRINT}%
{SON \MakeLowercase{\textit{et al.}}: PALETTESIM: FAST MULTI-ENGINE NEURAL SIMULATION}
\else
\author{Anonymous Author(s)}
\fi

\makeatletter
\IEEEaftertitletext{%
  \ifdefined\PaletteSimArxiv\vspace{-10.7mm}\else\vspace{-7pt}\fi%
  \begin{minipage}{\textwidth}
    \def\@captype{figure}
    \setlength{\abovecaptionskip}{2pt}
    \centering
    \includegraphics[width=\textwidth]{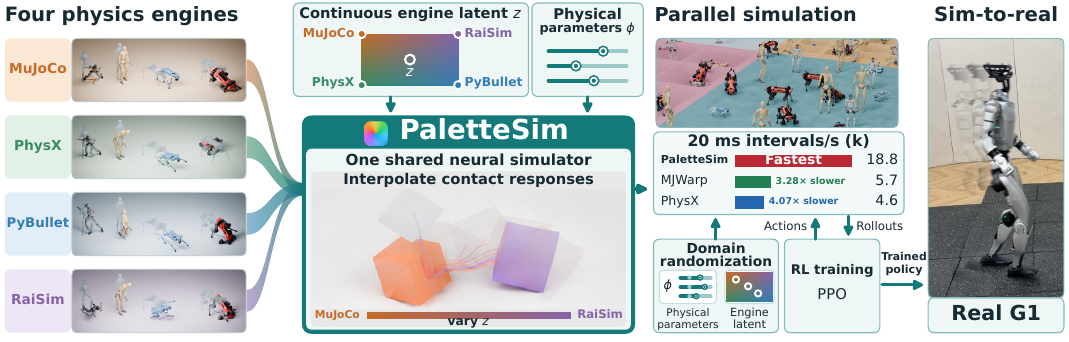}
    \caption{\textbf{Overview of PaletteSim.}
    Trajectories from four physics engines train one shared neural simulator.
    Its continuous engine latent $\mathbf z$ lets it interpolate between the
    engines' contact responses. Throughput reuses the RTX~5090 timing batch
    in Fig.~\ref{fig:geometry-rollout-speed}(d).
    We randomize $\mathbf z$ and the physical
    parameters $\phi$ during policy training. The resulting walking policy
    transfers to a real G1.}
    \label{fig:overview}
  \end{minipage}
  \vspace{0pt}%
}
\makeatother
\maketitle

\begin{abstract}
Robot learning needs fast simulators that capture real contact dynamics.
Detailed collision geometry slows analytical simulation, and a single
engine's contact model can miss real motion even when physical parameters
vary. We introduce PaletteSim, a neural simulator that learns contact
dynamics from four physics engines. Its continuous engine latent lets us
fit contact responses to measured motion and randomize them during policy
training. To accelerate each step, PaletteSim computes free motion
analytically and uses an $\mathrm{SO}(3)$-equivariant graph neural network
to predict point corrections for contacts and joint limits. PaletteSim achieves
higher throughput than MJWarp, PhysX, and learned baselines with reported
timings. After calibration, it matches recorded G1 walking and box-tossing
trajectories more closely than the evaluated baselines. A G1 walking policy
trained in PaletteSim also transfers to real hardware.
\ifdefined\PaletteSimArxiv
\href{https://dongwon-son.github.io/palettesim/}{\textcolor{blue}{Project website}}.
\else
See the supplementary material for videos.
\fi
\end{abstract}


\IEEEpeerreviewmaketitle

% Put the two introductory top floats in separate columns.
\setcounter{topnumber}{1}
\section{Introduction}

\begin{figure}[t]
  \centering
  \includegraphics[width=\columnwidth]{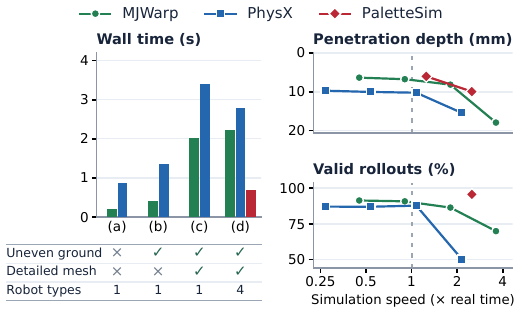}
  \caption{\textbf{Geometry cost and simulation stability.}
  \emph{Left} shows median wall time over five runs for 128 scenes simulated
  for 2\,s on RTX~5090. (a)--(c) use G1, and (d) uses four robot types.
  \emph{Right} shows stability across timesteps for 1,024 rollouts with
  PaletteSim, MJWarp, and PhysX. A rollout is valid if every sampled state
  is finite and within preset pose and speed bounds.}
  \label{fig:geometry-rollout-speed}
\end{figure}

To walk and manipulate objects reliably in the real world, robots need
extensive experience under varied \mbox{physical}
\mbox{conditions}~\cite{peng2018sim,muratore2022robot}. However, collecting
that experience on physical robots is slow, costly, and sometimes unsafe,
so training often relies on simulation~\cite{mujocoWarp,makoviychuk2021isaac}.
A simulator must therefore generate many trajectories quickly and capture
the range of motions observed in real robots and objects. Widely used
analytical physics engines face challenges in meeting both requirements.
First, detailed collision geometry slows simulation
(Fig.~\ref{fig:geometry-rollout-speed}), and using larger timesteps or
fewer solver iterations can compromise accuracy or
stability~\cite{son2023locc,erez2015simulation}. Second, each engine uses
the same contact rules across trials as physical parameters vary.
Because those rules simplify what happens at impact, adjusting the
parameters alone may fail to reproduce the observed
trajectory~\cite{acosta2022validating,fazeli2020fundamental,lelidec2024contact}.
Our goal is to simulate robots and objects quickly and to better match
real motion by modeling a wider range of contact responses.

Alternatively, learned simulators can generate trajectories quickly by
replacing costly collision and contact calculations with network
predictions~\cite{son2023locc,liao2026supermimic}. At each step, the
simulator takes the current poses and velocities, scene geometry, and any
robot commands as inputs and predicts the next state. Repeating the update with
each predicted state as the next input generates a trajectory. Providing
physical parameters as network inputs enables domain randomization
during policy training~\cite{peng2018sim}. However, a simulator trained on trajectories from
one physics engine sees only that engine's contact
responses~\cite{rubanova2024sdfsim,xu2025nerd}. If the engine mispredicts
an object's motion after impact, the simulator's learned dynamics model
may reproduce the same error.

% Saved measurements: figures/geometry_rollout_speed_20261106/FASTEST_TIMINGS.json.
% Defer the figure past the introduction text in the second column of page 1.
\suppressfloats[t]

% Overleaf thread 6afc6eae61552a784b000001
% Resolved in the preceding paragraph. \PaletteSimOverleafNote{Physical parameters as network inputs}{what about "to apply domain randomization with learned dynamics, ". Needs description of why users want to control the learned dynamics}

\begin{figure*}[!t]
  \centering
  \includegraphics[width=\textwidth]{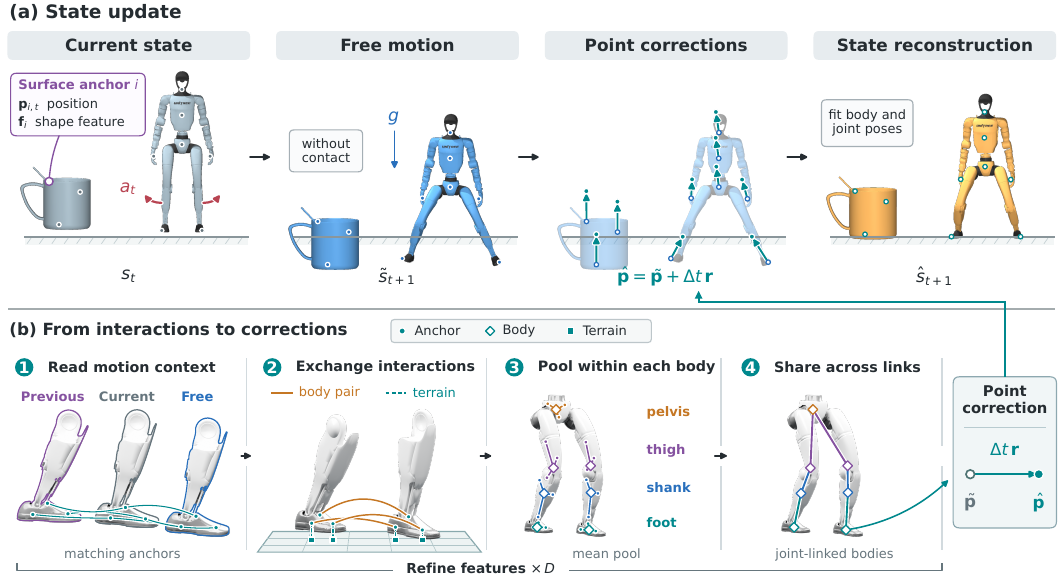}
  \caption{\textbf{How PaletteSim predicts the next state.}
  (a) Unitree G1 and mug examples illustrate free motion (blue),
  contact-resolved motion (orange), and point corrections (teal arrows).
  (b) Schematic of the point-correction network.}
  \label{fig:arch}
\end{figure*}

\begingroup
\setlength{\intextsep}{2pt}
\setlength{\columnsep}{5pt}
To speed up simulation and capture a wider range of contact responses
than a single engine provides, we introduce \emph{PaletteSim}, a neural simulator for articulated
robots, rigid objects, and terrain (Figs.~\ref{fig:overview}
and~\ref{fig:arch}). We train one network on trajectories from
MuJoCo~\cite{todorov2012mujoco}, PhysX~\cite{physxRigidDynamics},
PyBullet~\cite{coumans2021pybullet}, and RaiSim~\cite{hwangbo2018contact}.
A discrete engine indicator restricts the simulator to the four engines'
responses, and even the best of them can miss the observed
motion.{\parfillskip=0pt\par}
% Reserve the landing plot's height and spacing before starting the example.
\Needspace{7\baselineskip}
\begin{wrapfigure}{r}{0.435\columnwidth}
  \includegraphics[width=\linewidth]{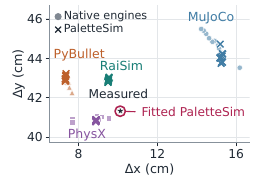}
\end{wrapfigure}
\noindent For example, in the box toss shown here, the closest predicted
landing point across all tested native engines and masses misses the
measured one by $7.0$\,mm. We therefore introduce the \emph{engine latent},
\mbox{a continuous input} that changes the simulated contact response.
We achieve this by sampling latent values from a separate region for each engine's
trajectories during training.
Interpolating between these regions lets PaletteSim
simulate intermediate contact responses, which can yield trajectories
closer to real motion. In this box toss, the fitted latent gives a
landing error of $0.10$\,mm. This flexibility supports contact-model identification
through latent optimization and extends domain randomization beyond
physical-parameter variation.\par
\endgroup
\medskip

\begin{figure}[t]
  \setlength{\abovecaptionskip}{2pt}
  \centering
  \includegraphics[width=\columnwidth,trim=0 3bp 0 0,clip]%
    {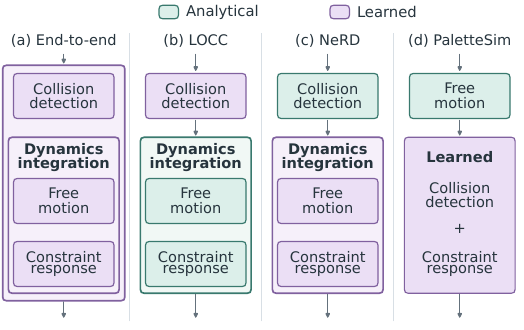}
  \caption{\textbf{Computation flow and learned components.}
  Arrows show information flow. Enclosures group functions handled together.
  Teal denotes analytical computation and violet denotes learned computation.
  Common inputs and final state reconstruction are omitted.}
  \label{fig:simulation-decomposition}
\end{figure}

To simulate these contact responses quickly over an entire trajectory,
we next consider which parts of each simulation step the network should
learn (Fig.~\ref{fig:simulation-decomposition}). An end-to-end model predicts
the entire next state directly from data (a). The same predictor must
learn how the system moves under applied forces (free motion), when bodies
come into contact (collision detection), and how constraints such as
contact change motion (constraint response). Abrupt changes in velocity
at impact can be particularly difficult to learn
accurately~\cite{parmar2021fundamental}, and the resulting prediction
errors can compound over successive steps~\cite{wei2025hopnet}.
To reduce this learning burden, hybrid methods compute part of the step
analytically. LOCC~\cite{son2023locc} learns only collision detection (b),
and NeRD~\cite{xu2025nerd} keeps analytical collision detection and learns
the remaining state update (c). These designs retain the cost of analytical
contact solving in LOCC~\cite{son2023locc} and collision checks on
detailed geometry in NeRD~\cite{xu2025nerd}.

These remaining costs motivate PaletteSim's design in
Fig.~\ref{fig:simulation-decomposition}(d). As in many rigid-body
engines, we first predict motion under inertia, gravity, and
actuation~\cite{mujocoComputation,physxRigidDynamics}. The fully
differentiable calculation skips costly collision checks and uses the
linear-time articulated-body algorithm for robots~\cite{featherstone2008rigid}.
A learned network then predicts \emph{point corrections}, which adjust the
free-motion positions of surface anchors to account for contacts and joint
limits. The network uses recent states
and controls, scene geometry, physical parameters, the engine latent,
and the free-motion state. We train the full prediction process on
trajectories generated by the four source physics engines.

To predict point corrections accurately with a compact network, we exploit
two properties of rigid-body dynamics. First, contact depends on the local
geometry where surfaces meet~\cite{son2023locc,son2025neuralsvcd}.
To exploit this locality, we represent each rigid body with surface
anchors~\cite{dou2026rigidformer} whose latent features $\mathbf{f}_i$ encode
the surrounding shape, then process them with a graph neural network
(GNN)~\cite{wu2021comprehensive} that limits contact messages to the closest
candidate pairs~\cite{rubanova2024sdfsim}. Second, point corrections obey
\emph{geometric symmetry}. Translating the entire scene leaves the
correction unchanged, while rotating the scene together with its velocities
and forces, including gravity, rotates the correction by the same rotation.
To exploit this symmetry, we use relative positions for translation
invariance and frequency-based equivariant features (FER)~\cite{son2024fer}
for $\mathrm{SO}(3)$ equivariance, so the network can learn a single contact
rule that applies across positions and orientations.

To simulate robots and objects quickly and better match their measured motion,
we make two contributions.
\begin{itemize}
  \item A continuous engine latent lets PaletteSim interpolate between
        the contact responses it learns from four physics engines to
        better match measured trajectories.
  \item To reduce computation while maintaining prediction accuracy, we
        split each simulation step into analytical free motion and learned
        point corrections. Our compact GNN exploits contact locality by
        limiting contact messages to selected pairs of surface anchors,
        and rotational symmetry by using FER to make the corrections
        $\mathrm{SO}(3)$-equivariant.
\end{itemize}

Across scenes with robots, objects, or both, PaletteSim achieves higher
throughput than the GPU-based engines MJWarp and PhysX at every tested
batch size, and it is faster and more accurate than the learned baselines
(Fig.~\ref{fig:throughput-google-convex} and Table~\ref{tab:rollout-surface}).
Removing free motion, FER, or the GNN each increases prediction error
(Table~\ref{tab:rollout-surface}). On held-out trajectories from real G1
walking and box-tossing trials, calibrated PaletteSim has the lowest
prediction loss among the evaluated simulators
(Fig.~\ref{fig:all-clip-sysid}(a)). Training on four engines lowers this
loss relative to a MuJoCo-only model, and fitting the continuous latent
lowers it further, by $20.4\%$ on G1 walking and $8.1\%$ on box tossing,
relative to the best fixed engine code (Fig.~\ref{fig:sysid-variants}).
Finally, a G1 walking policy trained in PaletteSim transfers to hardware
(Sec.~\ref{sec:exp-transfer}).
% Hardware claim is descriptive planar-path error from existing recordings,
% not higher steady tracking reward or an isolated engine-latent DR benefit.
% The multi-seed matched comparison remains planned.
\setcounter{topnumber}{2}

\section{Related Work}

\textit{Multi-engine training for control.}
PolySim~\cite{lei2025polysim} also studies the benefits of using multiple
simulators for humanoid control, a question we examine for G1 walking.
However, each rollout follows its source engine's contact model.
If none of the available models captures the real robot's
foot--ground contact dynamics, switching engines can still produce
trajectories that deviate from measured motion. On the other hand,
PaletteSim learns contact dynamics from four engines and conditions its
predictions on a continuous engine latent. This lets the same simulator
vary its contact response at fixed physical parameters. We use this
flexibility in system identification by fitting the latent alongside the
physical parameters. On held-out G1 walking and box-tossing data, the
fitted latent yields predictions closer to measured motion than the best
fixed engine setting
(Sec.~\ref{sec:exp-sysid}, Fig.~\ref{fig:sysid-variants}).

% Queue the architecture after Related Work starts, near the Method opening.
% Source-grounded PaletteSim inference/rollout overview.
% Vector PDF, editable SVG, and native-asset builder:
% figures/figure3_refined_20261107/native/.

\textit{Learned robot and object dynamics.}
In model-based RL, world models support policy training
(Dreamer~\cite{hafner2020dreamer}, SuperMimic~\cite{liao2026supermimic})
and action planning (TD-MPC~\cite{hansen2022tdmpc}).
These models are trained online on interactions collected by the policy
being optimized, and they learn the full state transition for that task.
As a result, a different policy can reach states and actions the model
has not seen. PaletteSim is instead trained offline on trajectories from
many controllers and scenes, and it learns only how contact modifies
analytically computed free motion. One trained model can therefore be
reused across tasks and policies.
To model contact interactions, FIGNet~\cite{allen2023learning} and
SDF-Sim~\cite{rubanova2024sdfsim} use graph neural networks over mesh
faces and surface points, respectively. LOCC~\cite{son2023locc} applies
3D convolutions to voxel grids, and Transformers attend across objects
in RigidFormer~\cite{dou2026rigidformer} and across time in
NeRD~\cite{xu2025nerd} and FlashNeRD~\cite{ataei2026flashnerd}.
PaletteSim follows RigidFormer in using surface anchors with local shape
features and SDF-Sim in limiting contact messages to a few pairs.
Unlike both, which predict the full update for rigid objects,
PaletteSim corrects analytical free
motion, adds joint--link edges, and uses
FER~\cite{son2024fer} for $\mathrm{SO}(3)$ equivariance.

\textit{Fitting contact dynamics to real motion.}
ContactNets~\cite{pfrommer2020contactnets} learns contact geometry from
measured trajectories, and Contact-Aware Neural Dynamics~\cite{jing2026contactaware}
predicts hand--object contact events and changes in object pose.
Huang et al.~\cite{huang2026fewshot} calibrate an analytical simulator to
real data before using it to generate training trajectories.
PaletteSim fits physical parameters and the engine latent directly to
measured trajectories while keeping the pretrained network fixed.

\section{Problem Definition}
\label{sec:problem}

We model the transition over one control interval as
\begin{equation}
  (\hat{s}_{t+1}, \hat{y}_{t+1}) =
  f_\theta(s_t, a_t, \Delta t, \mathcal G, h, \phi, \mathbf z),
  \label{eq:problem}
\end{equation}
where $s_t$ is the current physical state, $a_t\in\mathbb R^{n_a}$ gives
one target position for each of the $n_a$ actuated joints, and
$\Delta t\in\mathbb R_{>0}$ is the known control duration.
$\mathcal G$ specifies scene geometry and robot kinematics, and
$h\in\mathbb R^{H\times W}$ stores one height per terrain grid point.
The sizes of $s_t$ and the physical parameters $\phi$ depend on the
bodies and joints in $\mathcal G$. The simulator $f_\theta$ has learned
weights $\theta$. Its adjustable inputs are $\phi$ and the continuous
engine latent $\mathbf z\in\mathbb R^2$, which controls the learned
contact response.
It predicts the next state $\hat{s}_{t+1}$
and impulse magnitudes $\hat{y}_{t+1}$ for contacts with the ground and
between bodies. Each predicted state becomes the next step's input.
Internally, the simulator also carries surface-point positions and
features from one step to the next (Sec.~\ref{sec:correction-network}).

\textit{Given scene and initial state.}
A scene contains any number of rigid objects and at most one articulated
robot with rigid links. Object shapes are known. The robot comes from a
given asset set specifying link shapes, the kinematic tree, joint types,
axes and limits, collision rules, and whether the base is fixed or
floating. This geometry stays fixed during a rollout. The ground is
static and rigid, with a single height $h(x,y)$ at each horizontal
location $(x,y)$ on a grid of known bounds and spacing. This excludes
overhangs. The state includes the poses and linear and angular velocities
of the robot base and objects, plus joint positions $q$ and velocities~$\dot q$.

\textit{Adjustable inputs and modeling scope.}
The physical parameters $\phi$ include mass, inertia, friction,
restitution, joint stiffness and damping, motor inertia, and gravity.
Each source engine $e$ has a quadrant $Q_e$ (Fig.~\ref{fig:overview}).
Counterclockwise from the lower left, the order is
PhysX~\cite{physxRigidDynamics},
PyBullet~\cite{coumans2021pybullet}, RaiSim~\cite{hwangbo2018contact},
and MuJoCo~\cite{todorov2012mujoco}.
During training, we draw $\mathbf z$ from the density
\begin{equation}
  p(\mathbf z\mid e)=4\,\varphi_2(\mathbf z)\,\mathbf{1}\{\mathbf z\in Q_e\},
  \label{eq:engine-latent-density}
\end{equation}
where $\varphi_2$ is the standard bivariate Gaussian density and the
support is the full quadrant $Q_e$.
With $\theta$ fixed, we can choose $\phi$ and $\mathbf z$,
randomize them across rollouts, or fit
them to measured motion. Both stay fixed within a rollout. Interpolating
$\mathbf z$ can produce a contact response different from that of any
individual engine. Physical effects absent from all four training
configurations, such as body deformation and actuator backlash, lie
outside this scope.

\section{Method}
\label{sec:method}

During each control interval, PaletteSim computes where surface points
would move under gravity and the robot's commands. A single network pass
predicts point corrections that account for contacts and joint limits at
all substeps. PaletteSim fits poses and joint positions to the corrected
points and uses the final state to start the next interval
(Fig.~\ref{fig:arch}(a)). Repeating this update produces a simulated
trajectory, or \emph{rollout}.

\subsection{Rollout Overview}
\label{sec:rollout}
\label{sec:geometry}

\textit{Current state.} Each rigid object and robot link is a \emph{body}
with fixed surface points called \emph{anchors}. A shape encoder adapted
from TRELLIS.2~\cite{xiang2025trellis2} describes the surface around each
anchor $i$ with a feature vector $\mathbf{f}_i$. The body's pose gives
the anchor's world position $\mathbf{p}_{i,t}$ and rotates its shape
features with FER~\cite{son2024fer}. A convolutional network encodes
the fixed terrain height field $h$ once into a feature grid $\mathbf{F}_h$.
We typically use eight anchors per robot link and 16 per object. Box
tosses use the eight corners.

\textit{Free motion.} Each PaletteSim call advances the state by one
control interval $\Delta t$. Fitting the simulator can require matching
predicted joint angles to measurements taken between updates, and
rewards during policy training can depend on contacts within the interval.
We therefore divide $\Delta t$ into $S$ substeps of length $\Delta t/S$
and predict states and contacts at all $S$ substeps in one call. To
compute free motion at those same times, we integrate object accelerations
from rigid-body dynamics and robot accelerations from the articulated-body
algorithm (ABA)~\cite{featherstone1983articulated} over each substep. This
gives the free-motion states $\tilde{s}^{(1:S)}$ and anchor positions
$\tilde{\mathbf{p}}^{(1:S)}$ at the end of each substep.

\textit{Point corrections.} The network uses the scene inputs in
Eq.~\eqref{eq:problem}, the free-motion prediction, and the previous
anchor positions and features. We encode the engine latent $\mathbf z$
with a small network and add its output to each body's features, so
changing $\mathbf z$ changes the learned contact response. For anchor $i$
at substep $j$, we scale the network output $\mathbf{r}_i^{(j)}$ by
the full interval $\Delta t$, so $\mathbf{r}_i^{(j)}$ has units of velocity
and its scale does not depend on $\Delta t$. We add the result to the
free-motion position
\begin{equation}
  \hat{\mathbf{p}}_i^{(j)} = \tilde{\mathbf{p}}_i^{(j)}
  + \Delta t\,\mathbf{r}_i^{(j)}, \qquad j=1,\ldots,S.
  \label{eq:point-correction}
\end{equation}
The network also predicts impulse magnitudes for ground and body--body
contacts at each anchor and substep.

\textit{State reconstruction.} Each prediction provides $S$ sets of
corrected anchor positions. From these sets, we reconstruct $S$
configurations, one per substep, giving the robot's root pose and joint
positions and the poses of the rigid objects. We fit these configurations
to the corrected anchors using Kabsch alignment~\cite{kabsch1976best} for
each object and damped Gauss--Newton updates~\cite{wampler1986manipulator}
across all robot links. We estimate velocities and accelerations by finite
differences across substeps, using reconstructed configurations spaced by
$\Delta t/S$. The final configuration and its velocity, along with the
anchor features, carry over to the next control interval.

\subsection{Point-Correction Network}
\label{sec:correction-network}

Our graph neural network~\cite{sanchez2020learning} connects anchors,
bodies, and sampled terrain points. Each of its $D$ blocks updates
anchor features $\mathbf{f}_{i,t}$ in four steps (Fig.~\ref{fig:arch}(b)).

\textit{1) Read motion context.} Each current anchor receives information
from its matching anchors in the previous state and the free-motion prediction.

\textit{2) Exchange interactions.} Contact edges connect anchors on
different bodies or connect anchors to terrain points. A small network
maps node positions and features to vectors. We compute contact messages
only for the $k$ allowed pairs whose vectors are closest in Euclidean
distance, and weight the messages by their selection scores.

\textit{3) Pool within each body.} Each body node at its center of mass
averages the features of its anchors.

\textit{4) Share across links.} Each robot joint sends information to its
parent and child link nodes. A Transformer exchanges information among
all robot body nodes. Each anchor then receives its body's updated feature.

\textit{Node and message features.} Anchor inputs include the current
and previous positions and features and the free-motion anchor positions.
Terrain nodes sample the feature grid $\mathbf{F}_h$. Body inputs include
mass, static and dynamic friction, and restitution. Joint inputs include
stiffness, damping, armature, command error $a_t-q_t$, free-motion joint
displacement, and joint limits. Joint axes and gravity provide directions.
We normalize the physical parameters using training statistics and encode
them with small networks.
For free-motion, contact, and within-body messages, the connected nodes'
encoded inputs determine a scale and shift for each message feature
(FiLM~\cite{perez2018film}). Messages use relative positions and FER
features so the same interaction rule applies across translations and
rotations. Each block predicts point corrections and impulse magnitudes.
We supervise all blocks during training and use the final block for
Eq.~\eqref{eq:point-correction}.

\subsection{Data Generation and Training}
\label{sec:dataset}
\label{sec:training}

\textit{Data generation.} We train a single PaletteSim model on 251K
simulated trajectories spanning four physics engines. The data cover four
floating-base robots (G1, H2, Go2, and ANYmal-C) moving alone or with
objects on flat and uneven terrain, and a fixed-base FR3 arm manipulating
tabletop objects. We vary collision geometry and draw object meshes from
a library of roughly 8K assets from Google Scanned Objects, OmniObject3D,
and NOCS~\cite{downs2022google,wu2023omniobject3d,wang2019nocs}. The data
also include object drops, box tosses, and Kubric MOVi-A/B
scenes~\cite{greff2022kubric}.
Figure~\ref{fig:simulation-demos} illustrates several of these scene types.
We generate joint targets with motion-tracking controllers, including
OpenTrack and MimicLite~\cite{opentrack2026,mimiclite2026}, and add
Gaussian noise in some rollouts. The reference motions come from
LAFAN1~\cite{harvey2020robust} for humanoids and
Kine2Go~\cite{palucki2026kine2go} for quadrupeds.
FR3 runs a nonprehensile manipulation policy~\cite{cho2024corn}.

\textit{Multistep training.} Each trajectory's source engine determines
the region from which we sample the engine latent for its training
window. We start from a recorded state, replay any robot commands over
$L$ model calls, and feed each prediction into the next call. This
exposes PaletteSim to its own prediction errors during
training~\cite{ross2011dagger,talvitie2017selfcorrecting}. We use $L=20$
for robot-only trajectories, $L=8$ for box tossing, and $L=4$ for the
other scenes. Each call spans 20\,ms in our generated data and
$5/240$\,s in MOVi, with four substeps.
We minimize the multistep rollout loss~\cite{benechehab2024multistep}
\begin{equation}
  \mathcal{L}_{\mathrm{roll}} = \frac{1}{L}\sum_{t=1}^{L}
  \bigl(\ell_{\mathrm{pt},t} + \ell_{\mathrm{pose},t}
  + 0.01\,\ell_{\mathrm{vel},t} + 0.5\,\ell_{\mathrm{imp},t}\bigr).
  \label{eq:rollout-loss}
\end{equation}
The first three terms compare anchor displacements, changes in robot base
pose and joint positions, and end-of-step robot and object velocities
with recorded values. The impulse term $\ell_{\mathrm{imp},t}$ sums
the anchor impulse magnitude losses for ground and body--body contacts.
We average over valid steps and available targets and optimize with
AdamW~\cite{loshchilov2019adamw} at a learning rate of $3\times10^{-5}$ and weight decay of
$10^{-4}$, with gradient-norm clipping at $1$.

\section{Experiments}
\label{sec:experiments}
Our goal is to develop a fast simulator that can better match real
trajectories by varying its contact response over a continuous range.
To evaluate how well PaletteSim meets this goal, we first measure how
quickly it simulates robot and object scenes compared with analytical and
learned baselines (Secs.~\ref{sec:exp-throughput} and~\ref{sec:exp-rollout}).
We then compare calibrated predictions with held-out G1 walking and
box-toss recordings and test whether fitting the continuous engine latent
improves accuracy over the best fixed engine setting (Sec.~\ref{sec:exp-sysid}).
Finally, we evaluate sim-to-real transfer by comparing how well walking
policies trained in PaletteSim and MJWarp track the same velocity commands
on a real G1 (Sec.~\ref{sec:exp-transfer}).

We compare PaletteSim with analytical engines and the learned simulators
NeRD~\cite{xu2025nerd}, RigidFormer~\cite{dou2026rigidformer}, and
SDF-Sim~\cite{rubanova2024sdfsim}, where applicable.
Unless otherwise specified, we use domain-specific PaletteSim models
trained on their respective datasets. For PPO~\cite{schulman2017ppo} walking
policy training and all PaletteSim rollouts shown in the demonstration
videos, we use \emph{Unified} PaletteSim, a single model trained jointly
on the full dataset described in Sec.~\ref{sec:dataset}.

\subsection{Simulation Speed at Different Batch Sizes}
\label{sec:exp-throughput}

\begin{figure}[!t]
  \setlength{\abovecaptionskip}{2pt}
  \centering
  \includegraphics[width=\columnwidth]%
    {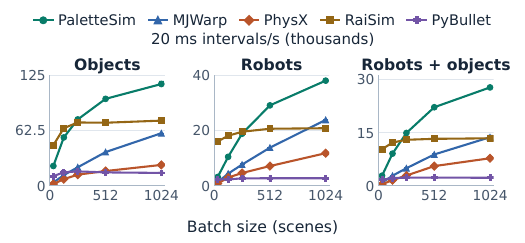}
  \caption{Simulation throughput on NVIDIA B200 GPUs and an AMD Ryzen~9
  9950X3D2 CPU. Throughput counts 20\,ms simulation intervals across all scenes
  per second of elapsed time.}
  \label{fig:throughput-google-convex}
\end{figure}

Figure~\ref{fig:throughput-google-convex} compares five simulators on batches
of 32 to 1,024 rough-terrain scenes. PaletteSim, MJWarp, and PhysX run on
B200 GPUs. RaiSim uses 32 CPU threads and \mbox{PyBullet} uses 24 workers.
For PhysX, we use the faster available result from OVPhysX or SAPIEN.
Scenes contain one robot, four objects, or both. Objects use
primitives or meshes from Google Scanned
Objects and OmniObject3D~\cite{downs2022google,wu2023omniobject3d}.
The robot mix has equal numbers of G1, H2, Go2, and ANYmal-C, driven by
Gaussian random commands. We use CoACD~\cite{wei2022coacd} convex colliders
with self-collision enabled. Each run simulates 2\,s at 5\,ms per substep,
with four substeps per PaletteSim call. Throughput uses the median of five
simulation-only timings after two warm-ups.

RaiSim leads at small batches. PaletteSim leads on all three workloads
at batches 512 and 1,024. At 1,024 robot scenes, it reaches about $38{,}100$ 20\,ms
intervals/s, $1.60\times$ MJWarp's throughput and $3.22\times$ PhysX's.
Moreover, PaletteSim's learned contact model predicts corrections for all
substeps in one pass, combining high speed with stable rollouts
(Fig.~\ref{fig:geometry-rollout-speed}). Analytical~engines have higher
estimated speeds at larger integration steps, with fewer valid rollouts
and deeper terrain penetration. With a 20\,ms update, PaletteSim
runs at $2.50\times$ real time for 128 scenes and yields more valid
rollouts ($95.7\%$) and shallower mean
terrain penetration than MJWarp ($69.9\%$ valid) and PhysX ($50.2\%$
valid) at 20\,ms integration steps. We next show that, after calibration,
PaletteSim also matches recorded motion more closely than the evaluated
baselines.

\subsection{System Identification}
\label{sec:exp-sysid}

System identification (SysID) fits the physical parameters $\phi$
listed in Sec.~\ref{sec:problem}, with one shared set for G1 walking
and another for box tossing. We keep PaletteSim's network weights fixed
during fitting. We use 40 non-overlapping G1 hardware clips from
Holosoma~\cite{holosoma} trials with random
joystick commands and from an in-house hurdle-crossing policy. We fit
the first 1.6\,s and replay commands for the next 0.5\,s. The G1 loss
combines squared joint-position, joint-velocity, and body angular-velocity
errors with the squared angle between body-frame gravity directions.
On the ContactNets box-toss dataset~\cite{pfrommer2020contactnets}, we fit
400 tosses and score 85 held-out tosses with weighted squared position and rotation errors.

\begin{figure}[tbp]
  \setlength{\abovecaptionskip}{2pt}
  \centering
  \includegraphics[width=\columnwidth]{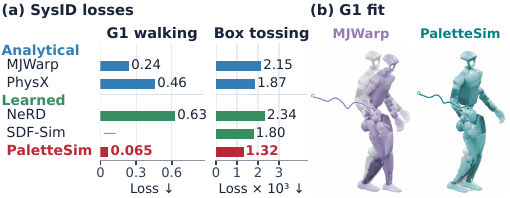}
  \caption{\textbf{System identification on real-world motion.}
  (a) Held-out prediction losses. (b) Simulated G1 paths over 1.6\,s and
  final poses, with recorded joints in gray.}
  \label{fig:all-clip-sysid}
\end{figure}

We use the cross-entropy method (CEM)~\cite{deboer2005tutorial} to fit the parameters, with limits of
80 iterations and 640 candidates per iteration.
PaletteSim jointly fits the physical parameters and one shared latent
$\mathbf z$ per task across all fitting clips, then reuses the fit for
held-out predictions.
In Fig.~\ref{fig:all-clip-sysid}(a), domain-specific PaletteSim has the
lowest loss on both tasks against MJWarp, PhysX, NeRD, and SDF-Sim (box
only). Figure~\ref{fig:all-clip-sysid}(b) shows the MJWarp and PaletteSim
fits for one G1 calibration clip.

\begin{figure}[tbp]
  \setlength{\abovecaptionskip}{2pt}
  \centering
  \includegraphics[width=\columnwidth]{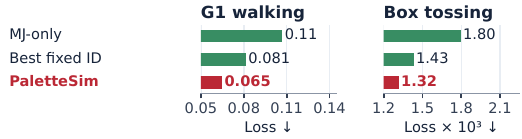}
  \caption{\textbf{PaletteSim variants after calibration.} Held-out
  prediction losses.}
  \label{fig:sysid-variants}
\end{figure}

Figure~\ref{fig:sysid-variants} compares three PaletteSim variants.
\emph{Best fixed ID} fits the physical parameters at one code per quadrant
and selects the code with the lowest fitting loss. The unit direction of
each code has its quadrant's signs and coordinate magnitudes of $1/\sqrt{2}$.
Its held-out loss is lower than that of \emph{MJ-only}, trained
only on MuJoCo data. Fitting the latent alongside physical parameters
reduces loss further by $20.4\%$ on G1 and $8.1\%$ on box tossing.

\subsection{Simulated Motion Prediction and Model Variants}
\label{sec:exp-rollout}

\begin{table}[!t]
  \centering
  \caption{\textbf{Held-out simulated motion prediction.} Errors are
  averaged over $0<t\leq1.2$\,s. Timing uses 128-scene batches on an
  RTX~5090. Bold marks the best value and underlining the second-lowest
  error.}
  \label{tab:rollout-surface}
  \footnotesize
  \setlength{\tabcolsep}{3pt}
  \renewcommand{\arraystretch}{1.08}
  \begin{tabular}{@{}lrrrr@{}}
    \toprule
    \textbf{Method} & \textbf{Robots} & \textbf{Robot--object} & \textbf{MOVi-A} & \textbf{MOVi-B} \\
    \midrule
    \multicolumn{5}{@{}l@{}}{\textit{Mean world-space surface-point error (cm) $\downarrow$}} \\
    % Full-geometry rough-terrain robot error receipts: notes/table_i_all_rough_robot_errors_20261119.json.
    % Latest matched accuracy refresh: notes/table_i_latest_refresh_20261120.json.
    % Continued PaletteSim checkpoints: notes/table_i_robot_continuation_20261120.json.
    % Verified robot-error improvements: notes/table_i_rolling_20261120.json.
    % Latest FlashNeRD re-evaluation: notes/table_i_flashnerd_54600_20261120.json.
    % MOVi-B uses the original SDF-Sim and PaletteSim scores.
    FlashNeRD~\cite{ataei2026flashnerd} & $34.99$ & -- & -- & -- \\
    RigidFormer~\cite{dou2026rigidformer} & $39.88$ & $28.96$ & $37.32$ & $33.55$ \\
    SDF-Sim~\cite{rubanova2024sdfsim} & -- & -- & $\underline{34.41}$ & $\underline{32.67}$ \\
    LOCC~\cite{son2023locc} & -- & -- & $43.65$ & $43.23$ \\
    \addlinespace[2pt]
    % Unified MOVi results from W&B checkpoint 1937161: notes/unified_movi_wandb_1937161_20261118.json.
    % Unified robot accuracy sources: notes/table_i_unified_robots_20261120.json.
    PaletteSim (Unified) & $\underline{17.47}$ & $\underline{15.70}$ & $36.25$ & $33.87$ \\
    PaletteSim (domain-specific) & $\mathbf{15.31}$ & $\mathbf{14.34}$ & $\mathbf{31.07}$ & $\mathbf{32.60}$ \\
    % Robot-domain ablation checkpoint selection: notes/table_i_ablation_robot_errors_20261119.json.
    % Robot-domain rechecks: notes/table_i_robot_continuation_20261120.json.
    % w/o FER final-checkpoint reevaluation: notes/table_i_no_fer_final_20261121.json.
    % MOVi ablation test-mean selection: notes/table_i_movi_ablation_best_20261121_1611.json.
    % Verified hourly ablation improvements: notes/table_i_hourly_best_20261122.json.
    \quad w/o FER & $65.93$ & $94.09$ & $37.25$ & $34.55$ \\
    \quad w/o free motion & $34.07$ & $34.99$ & $70.14$ & $48.15$ \\
    \quad GNN $\rightarrow$ Transformer & $60.48$ & $74.90$ & $70.77$ & $63.67$ \\
    \midrule
    \multicolumn{5}{@{}l@{}}{\textit{Throughput ($10^3$ forecast steps/s) $\uparrow$}} \\
    FlashNeRD & 7.97 & -- & -- & -- \\
    RigidFormer & 0.30 & 0.28 & 3.40 & 0.97 \\
    SDF-Sim & -- & -- & 0.50 & 0.55 \\
    LOCC & -- & -- & 14.71 & 14.70 \\
    \addlinespace[2pt]
    PaletteSim & \textbf{12.73} & \textbf{11.21} & \textbf{18.42} & \textbf{15.65} \\
    \bottomrule
  \end{tabular}
\end{table}

\begin{figure*}[!t]
  \setlength{\abovecaptionskip}{2pt}
  \centering
  \includegraphics[width=\textwidth,trim=0 4bp 0 0,clip]%
    {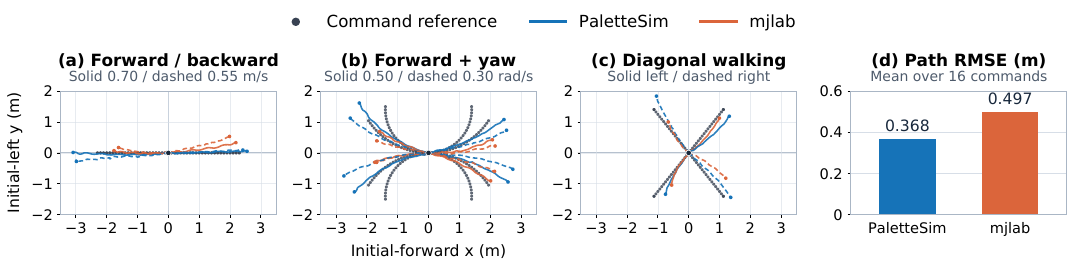}
  \caption{\textbf{Real G1 command-path tracking.} Paths share a starting
  position and heading. Panel (d) averages time-weighted planar position RMSE.}
  \label{fig:real-command-paths}
\end{figure*}

% Author confirmed original-motion exclusion, including Unified pretraining, Nov. 6, 2026.
To evaluate prediction accuracy, we compare PaletteSim with other learned
simulators and use ablations to assess its design choices.
Table~\ref{tab:rollout-surface} reports errors on four held-out simulation
datasets. The robot-only and robot--object sets each contain 1,024
trajectories of G1, H2, Go2, and
ANYmal-C on rough terrain with full collision geometry and self-collision.
We vary physical parameters such as mass and friction and generate the same
number of trajectories for each robot with each of the four physics engines.
The robot--object scenes add four rigid objects. We also evaluate on Kubric
MOVi-A and MOVi-B~\cite{greff2022kubric}, two benchmarks used in prior work
on rigid-body dynamics~\cite{allen2023learning,dou2026rigidformer}.
We hold out the original evaluation motions from every training dataset,
including the one used for Unified PaletteSim.

The baselines are the domain-specific learned models
FlashNeRD~\cite{ataei2026flashnerd},
RigidFormer~\cite{dou2026rigidformer}, SDF-Sim~\cite{rubanova2024sdfsim}, and
LOCC~\cite{son2023locc}, where applicable.
For the robot benchmarks, we train on trajectories pooled across the four
physics engines. Each method receives the recorded source-engine identity
during training and throughout each rollout.
We train three ablations for each task. One replaces the
$\mathrm{SO}(3)$-equivariant message layers with standard multilayer
perceptrons (w/o FER). Another removes the free-motion prediction from the
network inputs and takes current velocities, gravity, and any robot joint
commands directly (w/o free motion). The third replaces the GNN with a
Transformer over the whole scene (GNN $\rightarrow$ Transformer).
Each method starts from observed states and feeds its predictions into
subsequent steps, replaying recorded commands in robot scenes. We average
the Euclidean distance between corresponding world-space surface points
(64 per collision shape) over the first $1.2$\,s.

In Table~\ref{tab:rollout-surface}, domain-specific PaletteSim has the
lowest mean error and highest reported forecast throughput on all four
datasets. On robot-only scenes, it scores $15.3$\,cm compared with $39.9$\,cm for
RigidFormer and $35.0$\,cm for FlashNeRD, and reaches $1.6\times$
FlashNeRD's forecast throughput. On robot--object scenes, PaletteSim
scores $14.3$\,cm compared with $29.0$\,cm for RigidFormer.
All three ablations have higher mean error than the full model in every
domain. The Transformer variant gives the largest errors on both MOVi splits.
Removing FER gives the largest errors on robot-only and robot--object
scenes.

\begin{figure}[!htb]
  \setlength{\abovecaptionskip}{1pt}
  \centering
  \footnotesize
  \renewcommand{\arraystretch}{0.82}
  \begin{tabular}{@{}l@{\hspace{0.018\columnwidth}}l@{}}
    \textbf{(a) FR3 with CORN~\cite{cho2024corn}} &
    \textbf{(b) Drop of 15 objects} \\
    \includegraphics[width=0.41\columnwidth,trim=125bp 4bp 125bp 65bp,clip]{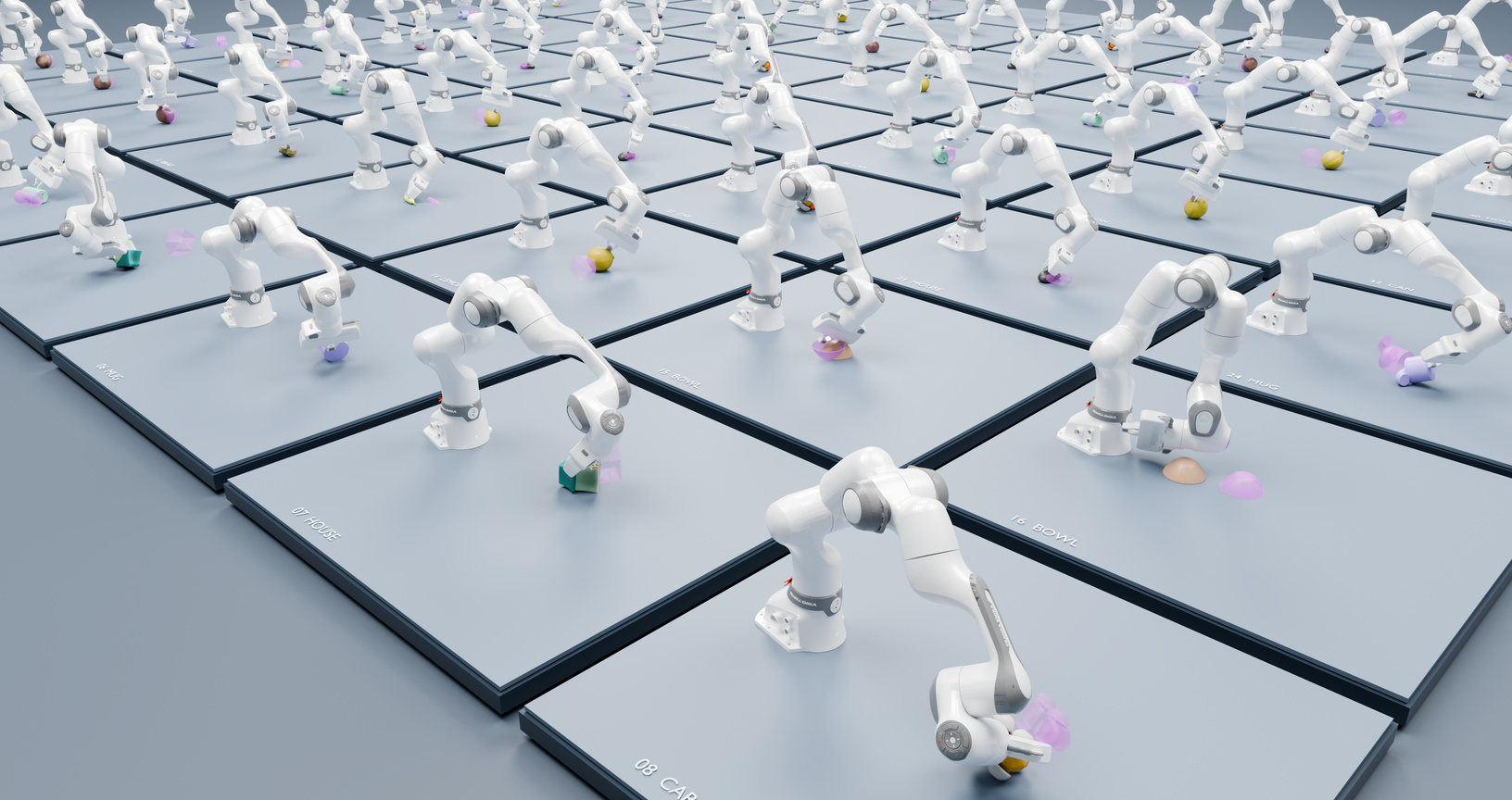} &
    \includegraphics[width=0.41\columnwidth,trim=0 28bp 0 28bp,clip]{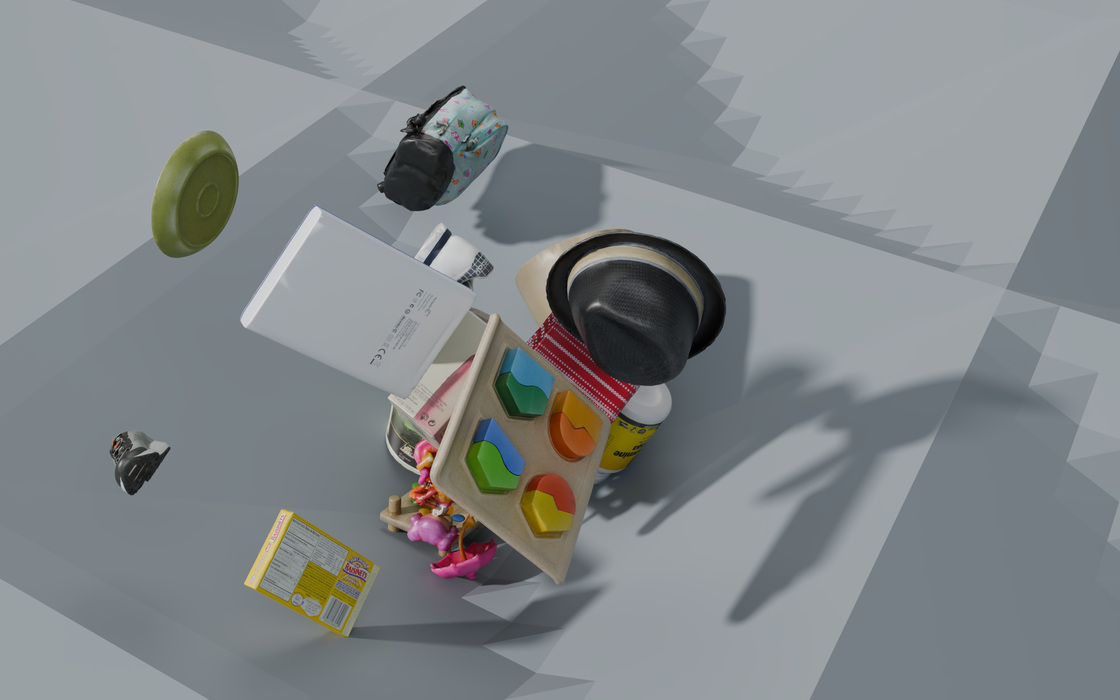} \\[1pt]
    \textbf{(c) Walking with objects} &
    \textbf{(d) Heterogeneous walking} \\
    \includegraphics[width=0.41\columnwidth,trim=0 18bp 0 38bp,clip]{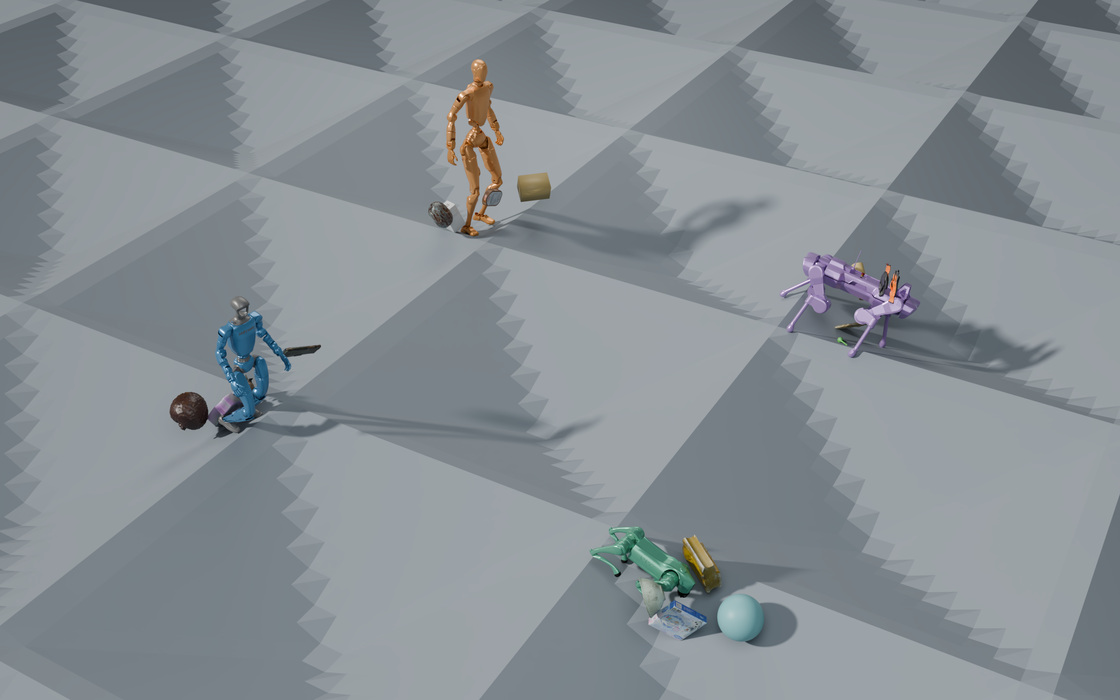} &
    \includegraphics[width=0.41\columnwidth,trim=0 6bp 0 50bp,clip]{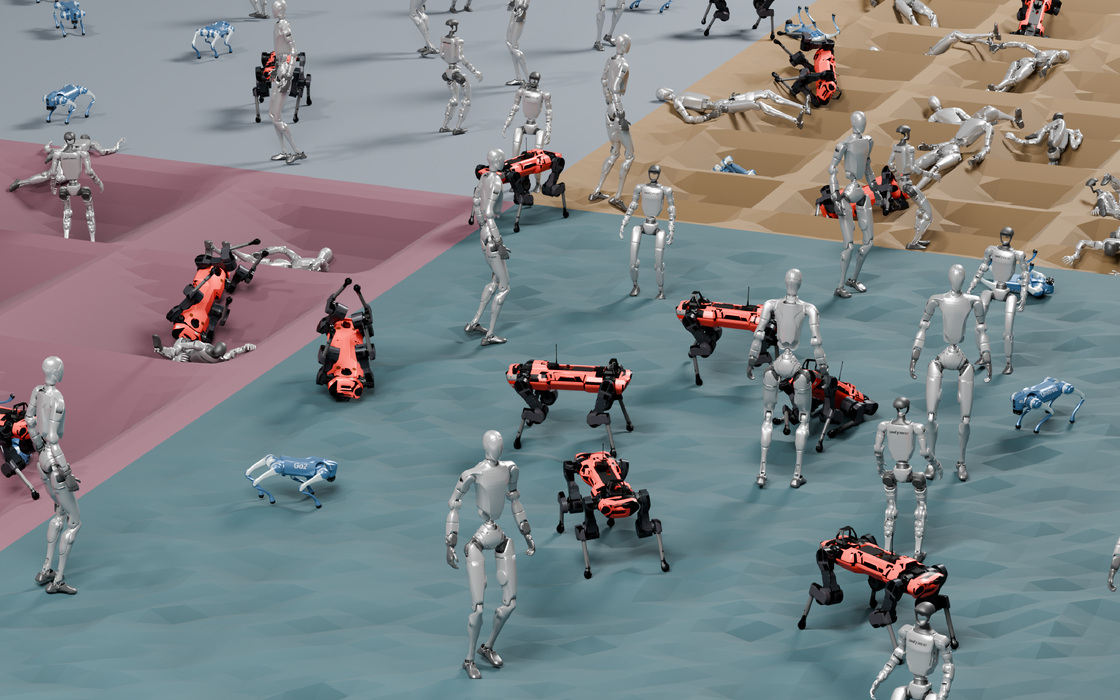}
  \end{tabular}
  \caption{\textbf{PaletteSim demos.}}
  \label{fig:simulation-demos}
\end{figure}

Unified PaletteSim uses one set of weights trained jointly on the full
dataset. Its mean errors are higher than those of the domain-specific
models on all four datasets (Table~\ref{tab:rollout-surface}).
Unified has lower error than every baseline on robot-only and robot--object
scenes and is within $2$\,cm of the best baseline on each MOVi dataset.
Unified PaletteSim can also simulate each scene type in
Fig.~\ref{fig:simulation-demos}.
The demo videos show more examples.

\subsection{Real-World Command-Path Tracking}
\label{sec:exp-transfer}

% Author confirmed matched policy-training settings on Nov. 6, 2026.
We train G1 walking policies with PPO in PaletteSim and MJWarp (through
mjlab~\cite{zakka2026mjlab}). Both use the same policy network, training configuration, reward
function, observation and action spaces, and randomization of physical
parameters. For PaletteSim, we also sample $\mathbf z\sim\mathcal N(\mathbf 0,I_2)$
at each PPO episode reset and hold it fixed throughout the episode, exposing
the policy to contact responses learned from multiple physics engines.
For the quantitative comparison, both policies receive the same recorded
sequence of body-velocity commands.
Figure~\ref{fig:real-command-paths}(a)--(c) compares the recorded and
commanded paths, and panel~(d) summarizes the errors.

We estimate pelvis motion from Vicon motion capture, the inertial
measurement unit (IMU), and joint measurements. We integrate the
commanded body-frame linear velocities and yaw rates to obtain the
reference paths. For each walking segment,
we align the paths once at the starting position and heading. We compute
the time-weighted root mean square error (RMSE) of $x$--$y$ position,
including command transitions, then average equally across the 16 settings.

PaletteSim has $25.9\%$ lower mean position RMSE than the MJWarp-trained
policy in these recorded trials (Fig.~\ref{fig:real-command-paths}(d)).
These results show that policies trained in PaletteSim transfer to
hardware.

% Saved stills and provenance are in figures/experiment_demos_20261106/.

\section{Conclusion}
\label{sec:conclusion}
We develop PaletteSim to simulate rigid-body dynamics quickly and match
real motion more closely. To adapt the contact response to measured
trajectories, we fit a continuous engine latent that interpolates among
responses learned from four physics engines. To speed up each step, we
compute free motion analytically and predict point corrections with
a graph neural network that exploits rotational symmetry through
$\mathrm{SO}(3)$-equivariant features. In our benchmarks, PaletteSim achieves
higher throughput than MJWarp and PhysX. Fitting the latent lowers held-out
prediction loss on measured G1 walking and box-tossing trajectories
relative to the best fixed engine code. The PaletteSim-trained policy
tracks commanded paths more accurately on a real G1 than the MJWarp-trained
policy. Currently, PaletteSim models rigid bodies on static height fields
with at most one articulated robot per scene.
To broaden this scope, we plan to model deformable bodies and
interactions between robots.

\bibliographystyle{IEEEtran}
\bibliography{references}

\end{document}